# WisenetMD: Motion Detection Using Dynamic Background Region Analysis

*Sang-Ha Lee, *Soon-Chul Kwon, **Jin-Wook Shim, **Jeong-Eun Lim, *Jisang Yoo

*Kwangwoon University, **Hanwha Techwin

**Abstract** – Motion detection algorithms that can be applied to surveillance cameras such as CCTV (Closed Circuit Television) have been studied extensively. Motion detection algorithm is mostly based on background subtraction. One main issue in this technique is that false positives of dynamic backgrounds such as wind shaking trees and flowing rivers might occur. In this paper, we proposed a method to search for dynamic background region by analyzing the video and removing false positives by re-checking false positives. The proposed method was evaluated based on CDnet 2012/2014 dataset obtained at "changedetection.net" site. We also compared its processing speed with other algorithms.

**Keywords** : motion detection, background subtraction, foreground segmentation, surveillance, video signal processing.

## 1) Introduction

Motion detection algorithms that can be applied to surveillance cameras such as CCTV (Closed Circuit Television) have been studied extensively. Motion detection algorithm is generally based on background subtraction by modeling background information and comparing the background with the current input image. There are many ways to model background information [1] [2]. In general, RGB color brightness value and features that can well represent background component are used. These features include Local Binary Pattern (LBP) [3], Local Binary Similarity Pattern (LBSP) [4], and so on. One research has revealed that LBSP has good performance when it is used with background subtraction algorithm [4]. In this paper, we proposed a method using background subtraction algorithm including RGB color value and LBSP [4] feature component. When using background subtraction algorithm, it is important to prevent false positives for dynamic backgrounds such as leaves and rivers. In previous research, we have defined and used parameters to prevent false detection in dynamic background. However, these methods do not adequately remove false positives. In this paper, we proposed a method to remove frequently occurring false positives. First, we defined dynamic background samples to collect false positive component. We generated a dynamic background region by analyzing the video scene. When the foreground is detected in the dynamic background region, it is removed by re-checking false positives in dynamic background samples. The proposed method was evaluated based on CDnet 2012/2014 dataset [5] [6] obtained from "changedetection.net" site. CDnet 2012/2014 dataset [5] [6] contains various environments such as camera jittering and scene that includes dynamic background. This paper described the proposed algorithm in Section 2, evaluated the performance and computation speed of the algorithm in Section 3, and concluded the paper in Section 4.

## 2) Method

The proposed method is a motion detection algorithm based on background subtraction. Flow chart of the method is shown in Fig. 1. The method consists of five modules: background samples, BG/FG classification, FP re-check, feedback process, and post-process. Background sample module is a module that saves and collects background samples used to classify between foreground and background. BG/FG classification module calculates the distance between the current input image and background samples to determine whether it is a foreground or a background. This is described in more details in Section 2.1 and Section 2.2. FP re-check module performs a re-check to prevent false positives that can occur in situations when backgrounds such as trees or rivers are moving. In this paper, we define a candidate region in which a dynamic background can exist and dynamic background samples to collect false positives. When the foreground is detected in the dynamic background region, it is removed by re-checking false positives in dynamic background samples. The feedback process module updates parameters used in background samples, BG/FG classification module. The post-process module is a module for processing filter operation and morphology operation to improve the quality of result image. This is described in more details in Sections 2.3, 2.4, and 2.5.

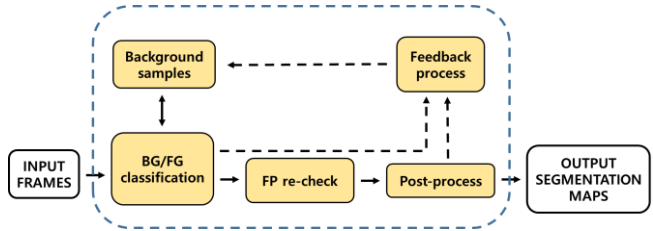

Figure 1) Flow chart showing the proposed method.

### 2.1) Background samples

Since the proposed algorithm operates based on background subtraction, it is essential to collect background samples. In this paper, we used sample consensus method used in SuBSENSE [7]. Background samples have the same resolution as the input image. The number of samples is fixed at 50 (=N).

$$B_t^n(x) \in \{\ B_t^1(x), B_t^2(x), \cdots, B_t^N(x)\ \} \tag{1}$$

Where t is index of frame, n is index of background samples, and x is index of pixel. Component of background samples consists of color and texture information. Color information uses RGB pixel value while texture information uses Local Binary Similarity Pattern (LBSP) [4] value. LBSP is a texture feature similar to LBP (Local Binary Pattern). Research has shown that using LBSP [4] feature in the background subtraction algorithm has good performance [8]. Mask shapes of LBP and LBSP are shown in Fig. 2.

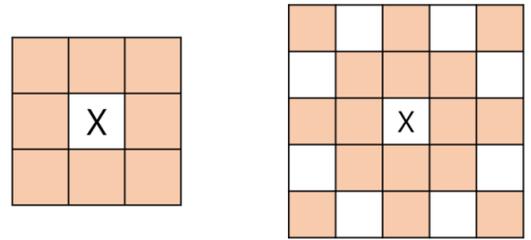

Figure 2) Mask shape of LBP (left) or LBSP (right).

LBP [3] uses pixel values of eight marked areas in a 3x3 mask while LBSP [4] uses pixel values of 16 marked areas in a 5x5 mask. Both texture features are calculated using pixel value of the marked area in the mask and the reference pixel value. In this paper, we used LBSP [4] calculation method used in SuBSENSE [7]. Equations (2) and (3) are used to calculate mask values of LBP [3] and LBSP [4], respectively.

$$m_{LBP}(i_p, i_x) = \begin{cases} 1 & if\ |i_p - i_x| \leq T_d \\ 0 & otherwise \end{cases} \tag{2}$$

$$m_{LBSP}(i_p, i_x) = \begin{cases} 1 & if\ |i_p - i_x| \leq T_r \cdot i_x \\ 0 & otherwise \end{cases} \tag{3}$$

In the above equation, $i_p$ is the pixel value of the area of the colored part in the mask, $i_x$ is the reference pixel value, $T_d$ is the threshold used in LBP [3], and $T_r(\approx 0.3)$ is the threshold ratio used in LBSP [4]. In Equation 3, LBSP [4] uses the reference pixel value and the threshold ratio to calculate the mask value. The mask value of LBSP [4] is calculated as 0 or 1 by comparing absolute difference of $i_p$ and $i_x$ with calculated threshold value. This method allows

LBSP feature values to adaptively respond to pixel distribution for the contrast. After calculating the value of the LBSP mask, the process of encoding these values into a 16-bit binary string is performed. Equation (4) represents a formula for converting calculated LBSP mask values into a 16-bit binary string.

$$LBSP(x) = \sum_{p=0}^{15} m_{LBSP}(i_p, i_x) \cdot 2^p \quad (4)$$

Since this process cannot individually store 16 binary numbers (1 bit) based on characteristics of computer structure, it is essential to have effective data capacity when expressed as one 16-bit number. Due to characteristics of a video, scenes will change as time passes. Therefore, it is essential that collected background samples are properly updated. In this study, background samples were updated using method of [1] [2]. For each frame, color component of the current input image and LBSP [4] component are collected with a probability of $1/T^t(x)$. Components to be updated are randomly selected and updated among 50 background samples. This update method has the advantage of using both previous and current information. The background update parameter $T^t(x)$ is calculated for each frame in the feedback process module.

## 2.2) BG/FG classification

The BG/FG classification module classifies foreground and background in the input image based on background sample information in the Background samples module. Equation (5) represents a formula for calculating foreground/background in the input image. This equation is the same as in SuBSENSE [7].

$$S^t(x) = \begin{cases} 1 & if \; N\{dist(I^t(x), B_n^t(x)) < R_{dist}^{t-1}(x)\} < 2 \\ 0 & otherwise \end{cases} \quad (5)$$

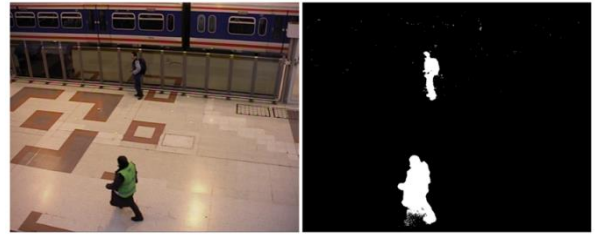

Figure 3) Input image $I^t(x)$ (left) and binary image $S^t(x)$ (right).

In the above equation, $I^t(x)$ is the input image while $S^t(x)$ is the binary image with foreground (1) and background (0) separated. $dist(I^t(x), B_n^t(x))$ returns L1 distance and hamming distance of the input image $I^t(x)$ and the background sample $B_n^t(x)$. Equation (6) shows how to calculate L1 distance. Fig. 4 shows how to calculate the hamming distance.

$$L1 \; distance = |I^t(x) - B_n^t(x)| \quad (6)$$

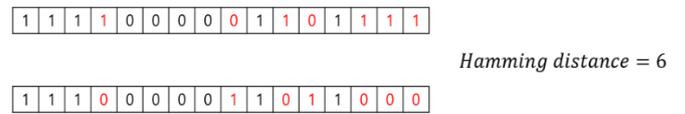

Figure 4) Examples of hamming distance calculation.

As shown in the above equation, L1 distance is a value obtained by calculating an absolute difference between two values. As shown in Fig. 4, the hamming distance is represented by the number of parts (red) with the same position but with different values for two arrays of the same length. Since LBSP [4] has 16 encoded binary arrays, hamming distance can efficiently represent the difference between two LBSP [4] values. $R_{dist}^{t-1}(x)$ returns color distance threshold $R_{color}^{t-1}(x)$ and LBSP distance threshold $R_{LBSP}^{t-1}(x)$. Equations (7) and (8) are formulas for calculating $R_{color}^{t-1}(x)$ and $R_{LBSP}^{t-1}(x)$.

$$R_{color}^t(x) = R_{color}^0 \cdot R^t(x) \quad (7)$$

$$R_{LBSP}^t(x) = 2^{R^t(x)} + R_{LBSP}^0 \quad (8)$$

In the above equation, $R^0_{color}$ and $R^0_{LBSP}$ are initial values of color distance threshold and LBSP distance threshold, respectively. $R^t(x)$ is a parameter that is updated in the feedback process. It is used to calculate the distance threshold. It is modeled to have a large value on a dynamic background and a value close to 1 on a normal background. It can prevent detection of false positives that can occur on dynamic backgrounds such as rivers and leaves.

### 2.3) FP re-check

FP re-check is a module that detects and removes false positives from dynamic background in binary image $S^t(x)$. In this paper, we defined a parameter representing the dynamic region. Equations (9), (10), and (11) are formulas for calculating the dynamic region.

$$DR^t(x) = \begin{cases} 1 & if\ BR^t(x) > blink\ threshold \\ 0 & otherwise \end{cases} \quad (9)$$

$$BR^t(x) = TB^t(x)/t \quad (10)$$

$$TB^t(x) = TB^{t-1}(x) + 1 \quad if\ S^t(x)\ xor\ S^{t-1}(x) = 1 \quad (11)$$

In ViBe + [2], the foreground and background tend to be periodically repeated in the dynamic background. This case is called "blinking". It can be expressed as the condition of Equation (11). In the above equation, $DR^t(x)$ is a parameter indicating the part with dynamic background in the input image, $BR^t(x)$ is the blinking rate per frame, and $TB^t(x)$ is the total number of blinking pixel. When a value of $BR^t(x)$ is higher than $blink\ threshold$, the pixel is regarded as a dynamic background region. Figure 5 shows the dynamic background region calculated from Equations (9), (10), and (11).

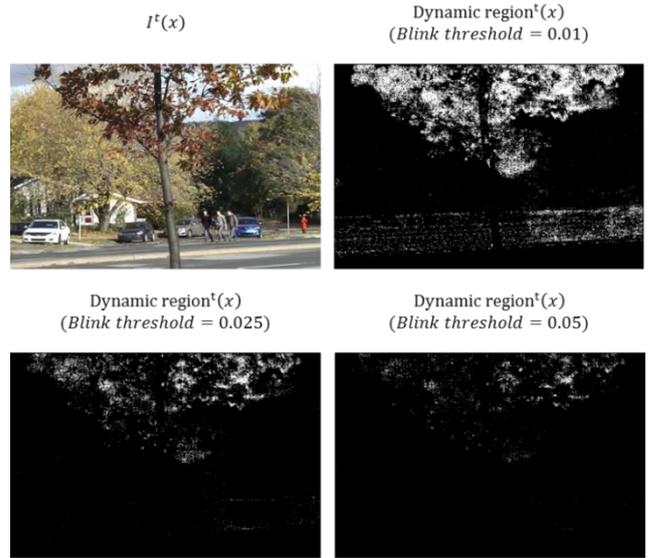

Figure 5) The dynamic background for blink threshold value.

The input image shown in Figure 5 is a video image in which a tree shakes in the wind and people and vehicles pass around. When the $blink\ threshold$ value is 0.01, the dynamic background is relatively well extracted. However, blinking occurs more frequently when the object passes. When $blink\ threshold$ is 0.05, the dynamic background is not extracted sufficiently. Experimental results showed that $blink\ threshold$ was 0.025.

We also defined dynamic background samples for collecting false positives. The dynamic background sample has the same resolution as the input image like background samples. The number of samples is fixed at 30. The dynamic background sample collects the color information of false detection component occurring in the dynamic background. Figure 6 represents conditions for collecting false positives.

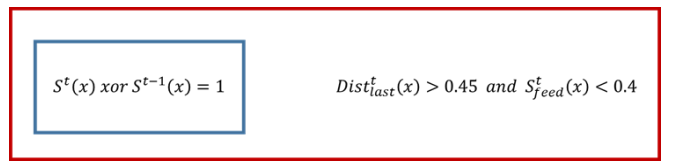

Figure 6) Conditions for collecting false positives.

The blue region in Fig 6 is a feature of the false positive component used in ViBe+ [2]. This indicates a case when previous result and the current result are different. If this happens frequently, it is likely to be a dynamic background in [2]. This is called "blinking". However, this feature tends to occur frequently even when object passes or when noise is severe. In this paper, we used additional conditions included in the red region to better represent the false positive component. Equations (12) and (13) show parameters used in Fig. 6.

$$Dist_{last}^t(x) = \left( \frac{L1dist(I^t(x), I^{t-1}(x))}{255 \times 3} + \frac{hdist(LBSP^t(x), LBSP^{t-1}(x))}{16 \times 3} \right) \div 2 \quad (12)$$

$$S_{feed}^t(x) = (1-\alpha) \cdot S_{feed}^{t-1}(x) + \frac{\alpha}{255} \cdot S^{t-1}(x) \quad (13)$$

In the above equation, $Dist_{last}^t(x)$ is a value obtained by normalizing the color/LBSP distance between the current frame and the previous frame to a range between [0,1]. $S_{feed}^t(x)$ is a parameter that expresses the trajectory of the object. It is composed of feedback of previous result $S^{t-1}(x)$. $Dist_{last}^t(x) > 0.45$ can prevent blinking due to image noise while $S_{feed}^t(x) < 0.4$ can prevent blinking caused by passing of an object through the image. When the condition of Fig. 6 is satisfied, the pixel is regarded as a false positive component and the color component is stored in dynamic background samples. This process selects and updates the sample randomly, like background sample update. When an object is detected on a dynamic background region, the color distance between the collected false positive and the pixel is calculated. If the value is smaller than the dynamic color threshold, it is regarded as background. This is because the false positive that occurs in the dynamic background tends to have distinct brightness differences from the background so that it is enough to distinguish false positives from color components alone. Through this process, it is possible to efficiently remove false positives that may occur in the dynamic background.

**2.4) Feedback process**

Feedback Process is a module that calculates $R^t(x)$ and $T^t(x)$ parameters used in Background samples and BG/FG classification module. Equations (14) and (15) are formulas used to calculate parameters.

$$D_{min}^t(x) = D_{min}^{t-1}(x) \cdot (1-\alpha) + d^t(x) \cdot \alpha \quad (14)$$

$$v^t(x) = \begin{cases} v^{t-1}(x) + w(x) & if\ S^t(x)\ xor\ S^{t-1}(x) = 1 \\ v^{t-1}(x) - v_{decr} & otherwise \end{cases} \quad (15)$$

$$w(x) = \begin{cases} 1.0 & if\ DR^t(x) = 0 \\ 1.5 & if\ Dist_{last}^t(x) > 0.45,\ S_{feed}^t(x) < 0.4 \\ 0.8 & otherwise \end{cases} \quad (16)$$

In the above equation, $\alpha$ is the learning rate and $d^t(x)$ is the minimum value of all color/LBSP distances normalized between the input image and the background sample. A smaller value of $d^t(x)$ is often a background while a larger value is a background or an object. $D_{min}^t(x)$ is fed back to $d^t(x)$ every frame. This feedback method increases the reliability of the value. We also used two learning rates to vary the feedback rate. These two values are 0.04 (short term) and 0.01 (long term). Short term contains more recent values while long term contains many older values. This update method allows the algorithm to be robust to the rate at which the environment changes.

$v^t(x)$ is a parameter that quantifies the degree of 'blinking' mentioned in FP re-check module. In this paper, we defined another parameter $w(x)$ to update this parameter. $w(x)$ has a large value in regions where there is a high probability of having a dynamic background. In this case, we used the condition to distinguish false positive components used in Fig. 6. Using this parameter allows $v^t(x)$ to increase to a larger value when blinking occurs in the dynamic background. Decreasing constant $v_{decr}$ uses a value of 0.1. Equations (17) and (18) are equations used for calculating $R^t(x)$ and $T^t(x)$ parameters.

$$R^t(x) = \begin{cases} R^{t-1}(x) + v^t(x) & if\ R^{t-1}(x) < (1 + D_{min}^t(x) \cdot 2)^2 \\ R^{t-1}(x) - \frac{1}{v^t(x)} & otherwise \end{cases} \quad (17)$$

$$T^t(x) = \begin{cases} T^{t-1}(x) + \frac{1}{v^t(x) \cdot D_{min}^t(x)} & if\ S^t(x) = 1 \\ T^{t-1}(x) - \frac{v^t(x)}{D_{min}^t(x)} & otherwise \end{cases} \quad (18)$$

The updating method of $R^t(x)$ and $T^t(x)$ is the same as that of SuBSENSE [7]. $R^t(x)$ is a parameter that is updated by $v^t(x)$. It has a high value for noisy or dynamic background. The condition of $R^{t-1}(x) < (1 + D_{min}^t(x) \cdot 2)^2$ is to have an exponential relationship when increasing the value of $R^t(x)$. This exponential relationship can reduce false positives by returning higher values of $R^t(x)$ for severely shaken trees.

$T^t(x)$ is a parameter used to update background samples. This parameter is used to determine whether to update background samples with a $1/T^t(x)$ probability every frame. In other words, the smaller the value, the more frequent updates will occur. It has a large value in the area where the object is detected while it has a small value in the area where the background is detected. It also has a small value within dynamic background. This is because the background is dynamic, thus requiring more color/LBSP components than a normal background.

## 2.5) Post process

Post process is a module that performs post-processing operations based on results obtained from the FP re-check module. To improve the quality of results, we proceeded with morphological operations and median filtering operations for resulting images. This process not only removes noise components in the resulting image, but also allows the foreground silhouette to be better represented.

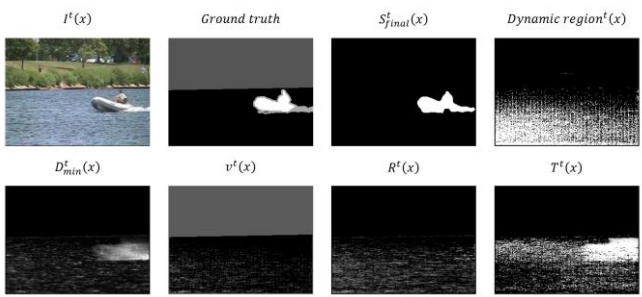

Figure 7) Parameter visualization used in the proposed algorithm.

## 3) Experimental results

In this paper, we evaluated and compared the proposed algorithm with changedetection.net. The site of changedetection.net provides dataset CDnet dataset for evaluating and comparing motion detection algorithms. CDnet dataset consists of two datasets: CDnet 2012 dataset and CDnet 2014 dataset. It includes various environments, ranging from sequences suitable for motion detection algorithms to sequences taken under harsh environments. In this paper, we evaluated the proposed algorithm with precision, FPR (False Positive Rate), FNR (False Negative Rate) for CDnet 2012/2014 dataset and compared it with other background subtraction algorithms.

## 3.1) Evaluation for CDnet dataset

CDnet 2012 dataset is basically object detection environment baseline, camera jitter, environment where moving background such as leaves, dynamic background, environment where motionless object exists, intermittent object motion, shadow environment. There are a total of six categories. Each category has four to six sequences. Table 1 compares other background subtraction algorithms with the proposed algorithm for CDnet 2012 dataset.

SuBSENSE [7] is one of these comparison algorithms. It is an algorithm that proposes a background modeling method using color/LBSP information based on ViBe+ [2] and PBAS [9]. PAWCS [10] proposes persistence of background samples based on SuBSENSE [7] algorithm. It has improved performance compared to SuBSENSE [7]. MBS (Multimode Background Subtraction) [11] is an algorithm that classifies backgrounds based on pixel sets using RGB and YCbCr channels. It classifies them using clustering. The proposed algorithm showed lower performance than PAWCS [10], but better performance than other algorithms.

Table 1) Comparison of different background subtraction algorithms and the proposed algorithm using CDnet 2012 dataset

| Method | Precision | FPR | FNR |
|---|---|---|---|
| **Proposed method** | **0.8650** | **0.0058** | **0.1584** |
| PAWCS [10] | 0.8746 | 0.0051 | 0.1453 |
| SuBSENSE [7] | 0.8576 | 0.0062 | 0.1719 |
| MBS [11] | 0.8480 | 0.0069 | 0.1897 |
| Spectral-360 [12] | 0.8461 | 0.0080 | 0.2230 |
| STBM [13] | 0.8210 | 0.0089 | 0.1650 |
| SGMM-SOD [14] | 0.8339 | 0.0062 | 0.2303 |

CDnet 2014 dataset includes 11 categories including bad weather, low frame rate, night video, camera zoom in/out and rotating PTZ, turbulence video, and six categories of CDnet 2012 dataset. Each of these categories also has 4 to 6 sequences. Table 2 compares other background subtraction algorithms with the proposed algorithm for CDnet 2014 dataset. WeSamBE [15] uses a method similar to PAWCS [10]. It sets weights on collected background samples and updates the background according to these weights. SharedModel [16] is a Gaussian Mixture Modeling (GMM) algorithm that takes the relationship between pixels and background samples into account.

The proposed algorithm showed lower performance than PAWCS [10] and WeSamBE [15] in terms of precision and FPR. However, the proposed method showed better performance than these algorithms in terms of FNR. Compared to SuBSENSE [7], it showed better performance.

Table 2) Comparison of different background subtraction algorithms and the proposed algorithm using CDnet 2014 dataset

| Method | Precision | FPR | FNR |
|---|---|---|---|
| **Proposed method** | **0.7668** | **0.0090** | **0.1802** |
| PAWCS [10] | 0.7857 | 0.0051 | 0.2282 |
| SuBSENSE [7] | 0.7509 | 0.0096 | 0.1876 |
| MBS [11] | 0.7382 | 0.0073 | 0.2611 |
| WeSamBE [15] | 0.7679 | 0.0076 | 0.2045 |
| SharedModel [16] | 0.7503 | 0.0088 | 0.1902 |

Table 3) Evaluation results of the proposed algorithm for CDnet 2012/2014 dataset

| | CD 2012 dataset | CD 2014 dataset |
|---|---|---|
| Average precision | 0.8650 | 0.7668 |
| Average FPR | 0.0058 | 0.0096 |
| Average FNR | 0.1584 | 0.1821 |

### 3.2) Processing speed

The proposed algorithm consists of C ++ language and OpenCV library. Testing for processing speed can measure fps at VGA resolution. The test environment consists of the following. It ran on a 7[th] generation Intel® Core™ i7 at 3.60GHz and GeForce GTX 1080 Graphics Card, using C++, CUDA, OpenCV. Table 4 compares the processing speed of the background subtraction method with the proposed algorithm in the above environment.

Table 4) Comparison of processing speed between the proposed method and other background subtraction algorithms

| Method | fps |
|---|---|
| **Proposed method** | **11~12(CPU), 33~35(GPU)** |
| PAWCS [10] | 2~3 |
| SuBSENSE [7] | 6~7 |
| WeSamBE [15] | 0.5~1 |

Table 5 shows comparison results between the proposed algorithm and other algorithms. The proposed method had a processing speed of 11 ~ 12 fps when only CPU was used. It had a processing speed of 33 ~ 35 fps when GPU was used. Among various algorithms used for comparison, PAWCS [10] showed the best performance, with processing speed of about 2 ~ 3 fps. The proposed algorithm was about 5 times faster. We also confirmed that the proposed algorithm showed better performance and fps than SuBSENSE [7] for CDnet 2012/2014 dataset.

### 4) Conclusion

In this paper, we defined a dynamic background region, searched for dynamic background, and newly defined dynamic background samples to collect false positive components that might occur in the dynamic background region, and then re-checked false positives. In this way, it is possible to remove false detection components that often occur in a dynamic background. However, since the dynamic background region of this paper is a parameter based on the assumption that "When blinking occurs, the region is likely to be a dynamic background region", there is a limitation in searching for a perfect dynamic background region. To improve this, it is necessary to find the unique tendency of the dynamic background and model it mathematically.